\documentclass[sigconf]{acmart} 
\usepackage{multirow}
\usepackage{booktabs}  

\newcommand{\eg}{\emph{e.g.,}}

\usepackage{siunitx} 
\usepackage{subfigure}
\usepackage{makecell}
\usepackage{pifont} 
\newcommand{\cmark}{\ding{51}} 
\newcommand{\xmark}{\ding{55}} 
\usepackage{enumitem} 
\setlist[enumerate]{leftmargin=*}
\usepackage{colortbl}
\definecolor{mygray}{gray}{.9}
\definecolor{mypink}{rgb}{.99,.91,.95}
\definecolor{mycyan}{cmyk}{.3,0,0,0}

\makeatletter
\renewcommand{\@thesubfigure}{\hskip\subfiglabelskip}
\AtBeginDocument{%
	\providecommand\BibTeX{{%
			\normalfont B\kern-0.5em{\scshape i\kern-0.25em b}\kern-0.8em\TeX}}}

\copyrightyear{2023}
\acmYear{2023}
\setcopyright{acmlicensed}\acmConference[MM '23]{Proceedings of the 31st ACM International Conference on Multimedia}{October 29-November 3, 2023}{Ottawa, ON, Canada}
\acmBooktitle{Proceedings of the 31st ACM International Conference on Multimedia (MM '23), October 29-November 3, 2023, Ottawa, ON, Canada}
\acmPrice{15.00}
\acmDOI{10.1145/3581783.3612299}
\acmISBN{979-8-4007-0108-5/23/10}

\begin{document}
	\title{Mutual Information-driven Triple Interaction Network for Efficient Image Dehazing}
%	\author{Hao Shen}
%	\affiliation{%
%		\institution{Hefei University of Technology, Hefei, China}
%		\city{}
%		\country{}}
%	\email{haoshenhs@gmail.com}
%	
%	\author{Zhong-Qiu Zhao}
%	\affiliation{%
%		\institution{Hefei University of Technology}
%		\country{}}
%	\additionalaffiliation{
%		\institution{Intelligent Interconnected Systems Laboratory of Anhui Province.}
%		\country{}
%	     }
%     \additionalaffiliation{
%     	\institution{Guangxi Academy of Sciences.}
%     	\country{}
%     }
% 	\authornote{Corresponding author.}
%	\email{z.zhao@hfut.edu.cn}
%	
%	\author{Yulun Zhang}
%	\affiliation{%
%		\institution{ETH Zürich, Zürich, Switzerland}
%		\country{}}
%	\email{yulun100@gmail.com}
%	
%	\author{Zhao Zhang}
%	\affiliation{%
%		\institution{Hefei University of Technology, Hefei, China}
%		\country{}}
%	\email{cszzhang@gmail.com}
%	+++++++++++++++++++++++++++++++++++++++++++++++++++++++++++++++++++++++++++++
	\author{Hao Shen}
	\affiliation{%
		\institution{Hefei University of Technology, Hefei, China}
		\city{}
		\country{}}
	\email{haoshenhs@gmail.com}
	
	\author{Zhong-Qiu Zhao}
	\authornote{Corresponding authors.}
	\affiliation{%
		\institution{Hefei University of Technology, Hefei, China}
		\country{}}
	\additionalaffiliation{
		\institution{Intelligent Interconnected Systems Laboratory of Anhui Province}
		\country{}
	}
	\additionalaffiliation{
		\institution{Guangxi Academy of Sciences}
		\country{}
	}
%	\authornote{Corresponding author.}
	\email{z.zhao@hfut.edu.cn}
	
	\author{Yulun Zhang}
	\affiliation{%
		\institution{ETH Zürich, Zürich, Switzerland}
		\country{}}
	\email{yulun100@gmail.com}
	
	\author{Zhao Zhang}
	\authornotemark[1]
	\affiliation{%
		\institution{Hefei University of Technology, Hefei, China}
		\country{}}
	\email{cszzhang@gmail.com}
	
	\begin{abstract}
		Multi-stage architectures have exhibited efficacy in image dehazing, which usually decomposes a challenging task into multiple more tractable sub-tasks and progressively estimates latent hazy-free images. Despite the remarkable progress, existing methods still suffer from the following shortcomings: (1) limited exploration of frequency domain information; (2) insufficient information interaction; (3) severe feature redundancy. To remedy these issues, we propose a novel Mutual Information-driven Triple interaction Network (MITNet) based on spatial-frequency dual domain information and two-stage architecture. To be specific, the first stage, named amplitude-guided haze removal, aims to recover the amplitude spectrum of the hazy images for haze removal. And the second stage, named phase-guided structure refined, devotes to learning the transformation and refinement of the phase spectrum. To facilitate the information exchange between two stages, an Adaptive Triple Interaction Module (ATIM) is developed to simultaneously aggregate cross-domain, cross-scale, and cross-stage features, where the fused features are further used to generate content-adaptive dynamic filters so that applying them to enhance global context representation. In addition, we impose the mutual information minimization constraint on paired scale encoder and decoder features from both stages. Such an operation can effectively reduce information redundancy and enhance cross-stage feature complementarity. Extensive experiments on multiple public datasets exhibit that our MITNet performs superior performance with lower model complexity. The code and models are available at \url{https://github.com/it-hao/MITNet}.
	\end{abstract}
	
	
	
	\maketitle
	\vspace{-0.2cm}
	\section{Introduction} \label{sec:introduction}
	Hazy images usually cause the degradation of visual quality~\cite{tan2008visibility,gui2023comprehensive} in object appearance, contrast, and color distortion, leading to significant performance drops in subsequent high-level vision tasks such as scene understanding~\cite{sakaridis2018model}, semantic segmentation~\cite{ren2018deep}, and object detection~\cite{liu2018improved}. Since these tasks urgently require clean haze-free images, academic and industry communities, thereby, have focused on single image dehazing, which attempts to extract latent clean images from hazy ones.
	
	Benefit from the important breakthrough of deep learning in computer vision tasks~\cite{tang2020blockmix,tang2022learning,zha2023boosting,li2023knowledge,yan2022image,wang2022fineformer}, data-driven image restoration~\cite{chen2023learning,chen2023hybrid,zhao2022fcl,zhang2022deep} and image dehazing methods~\cite{ffa_net,msbdn,aecr_net,fsdgn,dehamer,maxim,chen2022unpaired,cong2020discrete} have achieved superior performance. For the image dehazing task, one can divide these methods into two categories: physics-free methods and physics-aware methods. The former first calculates the transmission map and atmospheric light independently before employing the atmospheric scattering model (ASM)~\cite{asm} to produce haze-free images. The latter tries to explore a mapping from hazy images to clean counterparts directly in an end-to-end manner. Among them, multi-stage architectures~\cite{mprnet,hinet,gfn,fsdgn} play a significant role in performance improvement. The basic idea is to decompose a challenging task into multiple easy sub-tasks and adopt progressive learning. Despite the remarkable advancement, there still exist some issues. 
	
	(1) \textbf{\textit{Limited exploration of frequency domain information.}} Most existing image dehazing methods~\cite{mprnet,hinet,gfn,pmdnet} focus exclusively on exploiting spatial features but fail to sufficiently leverage frequency discrepancies, such as the intrinsic prior of the physical properties of hazy images, thus limiting performance gains. As revealed in~\cite{fsdgn}, the degraded haze information is almost displayed in the amplitude component, while the disparity between phase components of the corresponding hazy and clean images is small. Given this discovery, how to embed frequency domain prior information into spatial features and effectively incorporate them into a unified dehazing network is an open issue. 
	
	(2) \textbf{\textit{Insufficient information interaction.}} Current multi-stage architectures mainly consist of ``early fusion'' and ``cross-stage fusion'' designs. The first category~\cite{gfn} attempts to directly fuse the previous stage's output and the current stage's input at the beginning layer, which ignores the propagation of intermediate features from earlier to later stages. Thus, the complementary information will be gradually weakened with the increasing network depths. The second category~\cite{fsdgn,hinet} fully integrates the contextually-enriched features from one stage to the next, easing the information loss caused by the encoder-decoder's repetitive up- and down-sampling operations. However, the fusion scheme only considers the interaction between paired scale features of two stages, ignores the exchange of cross-scale information of in-stage and cross-stage, and thereby fails to provide more precise and contextually enriched feature representations. Besides, recent research~\cite{cran,mallnet,cgc,adfnet} indicates the function of neurons should be modified adaptively in response to behavioral context. Thereby, it is vital to modify the convolution operation based on context information dynamically.
	
\begin{figure}[!t]
		\centering
		\includegraphics[width=0.44\textwidth]{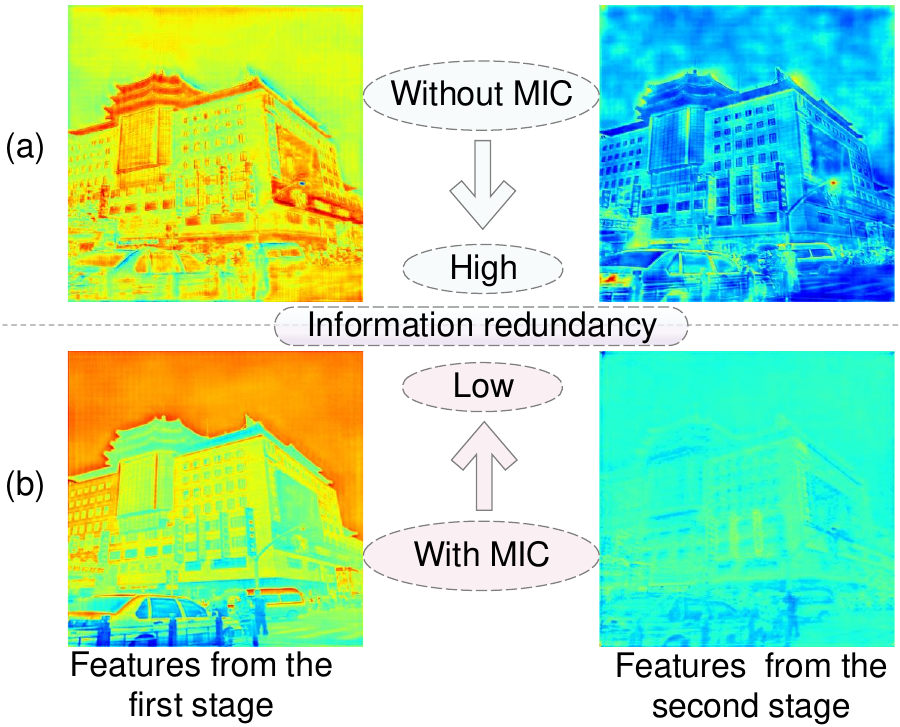}
		\vspace{-2mm}
		\caption{The feature comparisons before and after applying mutual information constraint (MIC).}
		\vspace{-2mm}
		\label{fig:motivation2} 
	\end{figure}

	(3) \textbf{\textit{Severe feature redundancy.}} Although performance can be improved by effective fusion strategies, existing dehazing methods do not explicitly enforce complementary information learning between different stage features, resulting in redundant features. We can observe from Figure~\ref{fig:motivation2} (a) that the learned features from the two stages have similar textures without any constraints. Conversely, information redundancy is reduced after endowing the mutual information constraint for two-stage features, as shown in Figure~\ref{fig:motivation2} (b). Thus, it is desirable to enforce complementary feature learning by adopting explicit constrain.
	
	In this work, firstly, a simple yet effective two-stage network is designed by simultaneously processing the spatial-frequency dual domain information. The first stage is designed to recover the amplitude spectrum of the hazy images for haze removal, and the second stage learns the transformation and refinement of the phase spectrum. To implement the network, we combine spatial-frequency information to construct the customized feature extraction block,~\eg~residual amplitude block for the first stage and residual phase block for the second stage, respectively. 
	Secondly, we propose an adaptive triple interaction module (ATIM) to enhance information exchange. Specifically, we design a triple interaction solution to achieve cross-domain, cross-scale, and cross-stage feature integration, and then an adaptive dynamic filter block is developed to generate dynamic filters based on the fused features. The produced filters next are convolved with the decoder features of the second stage for representation capability enhancement. Since the generated filters are conditioned on the input features guided by Fourier prior and spatial information, the network can flexibly adapt to image contents. 
	After being equipped with the two-stage architecture and the proposed ATIM, diverse frequency domain information, and rich spatial contextual information are exploited. To learn complementary information and alleviate feature redundancy from two stages, we introduce the mutual information minimization constraints on paired scale encoder and decoder features from two stages.
	Overall, the main contributions of this paper can conclude as follows:
	\begin{enumerate}
		\item We propose a novel dehazing network termed MITNet, based on spatial-frequency dual domain information and two-stage architecture, which simplifies the challenging dehazing problem into two more manageable sub-tasks and embraces the advantages of high performance and lower model complexity.
		\item We design an adaptive triple interaction module (ATIM) capable of aggregating cross-domain, cross-scale, and cross-stage features and generating content-adaptive filters to enhance global context representation.
		\item We introduce the mutual information minimization constraint on the paired scale encoder and decoder features from two stages to reduce information redundancy and enhance cross-stage feature complementarity.
		\item Comprehensive experiments on several public benchmarks verify the superiority and strong generalization. Further, our method achieves an excellent model complexity versus performance trade-off.
	\end{enumerate}
	\begin{figure*}[t]
		\centering
		\includegraphics[width=0.85\textwidth]{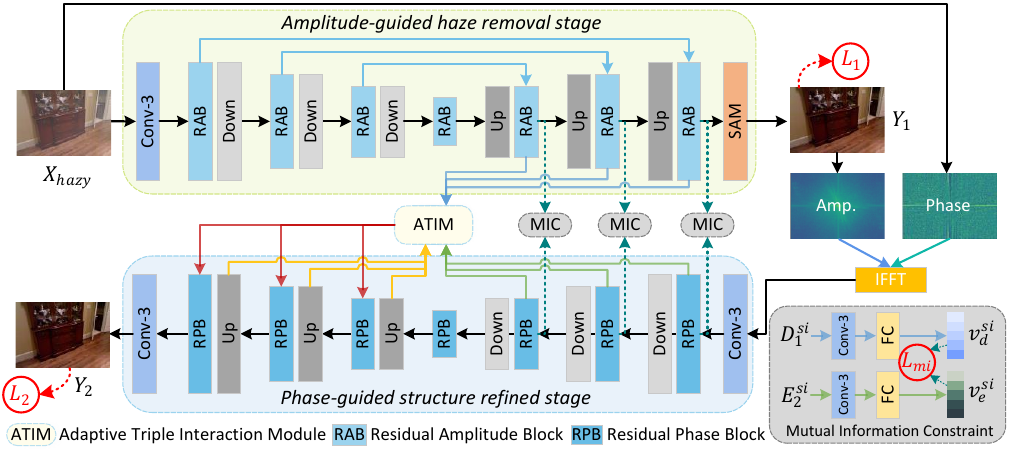}
		\vspace{-2mm}
		\caption{The detailed architecture of our Mutual Information-driven Triple interaction Network.}
		\vspace{-2mm}
		\label{fig:net} 
	\end{figure*}
	\section{Related Work}
	\subsection{Single Image Dehazing} Existing single image dehazing methods can be classified into prior-based and data-driven deep learning methods. The former~\cite{dcp, cap, nlp,mrf,Zhu2014SingleID,grm,zhang2017fast} usually rely on the ASM or handcraft priors. However, these priors only work well in the scene that satisfies their assumptions. 
	
	With the introduction of sizeable hazy datasets~\cite{reside,haze4k}, a plethora of data-driven approaches have been proposed rapidly in the past years. Early methods~\cite{dehazenet, aod_net} usually design a network to estimate the transmission map and atmospheric light. However, inaccurate prediction for the atmospheric light and transmission map may lead to performance degradation. Recent research~\cite{qu2019enhanced,gfn,gdn,msbdn,ffa_net,aecr_net,dehamer,fsdgn, ridcp} has concentrated on recovering the clean image directly from the hazy image without the need for a physical model. GFN~\cite{gfn} proposes a revolutionary fusion-based strategy for generating haze-free images. FFA-Net~\cite{ffa_net} introduces a channel attention and pixel attention mechanism for constructing a feature attention block that can handle many forms of input. MSBDN~\cite{msbdn} utilizes boosting and error feedback mechanisms to enhance the feature fusion. AECR-Net~\cite{aecr_net} exploits negative and positive image information via contrastive learning, achieving great network parameters versus performance trade-offs. Dehamer~\cite{dehamer} combines CNN and Transformer to fuse local representation and global context modelling capability perfectly to achieve haze removal. However, these methods only utilize the spatial domain features but fail to take advantage of the frequency domain information. FSDGN~\cite{fsdgn} uses spatial-frequency information to achieve fast inference speed but fails to achieve the complementarity of cross-stage features.
	
	In addition, current mainstream image restoration methods, including image dehazing~\cite{fsdgn}, image denoising~\cite{adfnet}, image inpainting~\cite{zheng2022deep,zheng2021gcm}, and image deblurring~\cite{fcl-gan} adopt U-Net~\cite{unet} architecture to pursue the model's efficiency. However, these methods neglect the interaction of cross-scale features. In contrast, full-resolution methods~\cite{ffa_net,jsnet} can obtain fine-grained features but fail to exploit contextual information. This paper inherits the benefits of U-Net and full-resolution architectures to develop a two-stage network.
	
	\subsection{Mutual Information Learning} 
	Mutual information has deep connections with representation learning~\cite{bengio2013representation}, aiming to capture the most important features similar to the input ideally. By extracting global features from the full image and local features from image patches, DIM~\cite{dim} estimates and maximizes mutual information. CPC~\cite{cpc} enables mutual information maximization between the context and features extracted from different sequential data components. CMC~\cite{cmc} employs contrastive learning in a multiview context, seeking to maximize mutual information across representations of different views of the same scene. Based on the empirical of multiview features, Info3d~\cite{info3d} enables mutual information maximization between 3D objects and the corresponding geometric converted equivalents based on empirical multiview features to improve representations. CMINet~\cite{cminet} specifically uses multi-modal features in color images and depth data through mutual information learning, removing redundant features for effective multi-modal learning. To learn mutual information on diverse image modalities, \cite{zhou2022mutual} implements mutual information minimization across PAN and MS features. Our proposed two-stage network explores different frequency domain characteristics in each stage and extracts rich spatial contextual features in both stages. Thus, common and diverse features exist in separate stages. To this end, we also apply mutual information as a constraint to alleviate the feature redundancy. 
	
	\section{Methods}
	\subsection{Network Architecture}	
	It is well known that the degradation information in low-quality images predominantly displays in the amplitude spectrum. Thus, we utilize the insight wisely and disentangle the dehazing procedure into a two-stage process so as to achieve haze degradation removal and preserve image structure information. Next, we will go through the operation and properties of the Fourier transformation. 
	
	\begin{figure*}[htbp]
		\centering
		\includegraphics[width=0.85\linewidth]{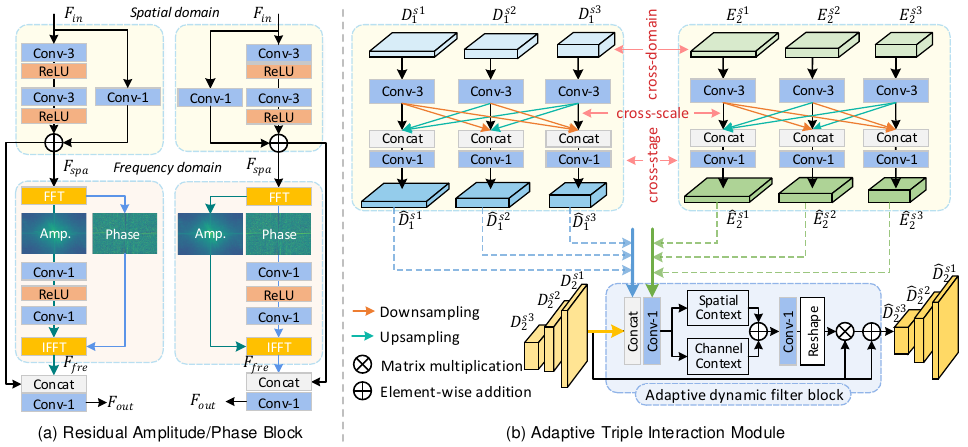}				
		\vspace{-2mm}
		\caption{The detailed architecture of (a) residual amplitude/phase block and (b) adaptive triple interaction module.}
		\vspace{-2mm}
		\label{fig:block} 
	\end{figure*}
 
	For a image $X_{img} \in R^{H\times W \times 1}$ with single channel, its Fourier transformation operation~\cite{fsdgn} $\mathcal{F}$ can be expressed as:
	\begin{equation}
		\mathcal{F}(X_{img})(i, j)=\frac{1}{\sqrt{H W}} \sum_{h=0}^{H-1} \sum_{w=0}^{W-1} X_{img}(h, w) e^{-k 2 \pi\left(\frac{h}{H} i+\frac{w}{W} j\right)}.
	\end{equation}
	Accordingly, $\mathcal{F}^{-1}$ is the inverse Fourier transformation, $k$ represents the imaginary unit, $i$ and $j$ denote the horizontal and vertical coordinates. Following with FFT algorithm~\cite{fouier}, the amplitude $\mathcal{A}(X_{img})$ and phase $\mathcal{P}(X_{img})$ information can be defined as:                    
	\begin{equation}
		\begin{aligned}
			& \mathcal{A}(X_{img})(i, j))=\sqrt{Real^2(X_{img})(i, j) +Imag^2(X_{img})(i, j)}, \\
			& \mathcal{P}(X_{img})(i, j))=\arctan \left[\frac{Imag(X_{img})(i, j)}{Real(X_{img})(i, j)}\right],
		\end{aligned}
	\end{equation}
	where ${Imag}(X_{img})$ and ${Real}(X_{img})$ denote the imaginary and real component of $\mathcal{F}(X_{img})$. Here, it is worth emphasizing that the Fourier operation is computed alone in each channel for feature maps or color images.
	
	The entire network, as shown in Figure~\ref{fig:net}, contains an \textbf{\textit{amplitude-guided haze removal stage}} and a \textbf{\textit{phase-guided structure refined stage}}. The former restores the amplitude of hazy images to remove haze, and the latter restores the phase information to refine fine-grained background structures. Let $X_{hazy}$ and $X_{gt}$ denote hazy and ground truth images, $Y_{1}$ and $Y_{2}$ represent the outputs of two stages. Since each stage is designed by a task-specific sub-network, both the input and supervision signals should be different. For the first stage, the hazy images $X_{hazy}$ are served as input, $\mathcal{F}^{-1}(\mathcal{A}(X_{gt}), \mathcal{P}(X_{hazy}))$ and $\mathcal{A}(X_{gt})$ are used to supervise the learning of amplitude representation. For the second stage, the $X_{gt}$ are used as the supervision signal, but we do not directly use $Y_{1}$ as the input of the second stage. Instead, we use $\mathcal{F}^{-1}(\mathcal{A}(Y_{1}), \mathcal{P}(X_{hazy}))$ to ensure the retention of original phase information.
	
    To achieve an efficient target, we use encoder-decoder-like (or U-Net-like)~\cite{unet} design for each stage, consisting of seven basic units (~\eg~residual amplitude block (RAB) and residual phase block (RPB)), three downsampling blocks, and three upsampling blocks in each stage. In the first stage, except for the lowest scale, each scale contains a skip connection between the encoder and decoder. However, considering that the decoder of the second stage integrates features from the previous stage and the current encoder by adopting the proposed fusion module, we do not take a similar approach. In addition, we introduce a supervised attention module (SAM)~\cite{mprnet} to stabilize the optimization procedure.
	
	We show the structure of RAB and RPB in Figure~\ref{fig:block} (a). Taking the former as an instance, which has a spatial domain branch and a frequency domain branch for processing dual domain representations. The spatial domain branch employs a residual block with two $3\times3$ convolutional layers to process spatial information. After obtaining the spatial features $F_{spa}$, we firstly utilize the fast Fourier transformation (FFT) to obtain the amplitude spectrum $\mathcal{A}(F_{spa})$ and phase spectrum $\mathcal{P}(F_{spa})$ information. Then, the $\mathcal{A}(F_{spa})$ is fed into two $1\times1$ convolutional layers to obtain $\mathcal{A}^{'}(F_{spa})$. Subsequently, we use the inverse FFT (IFFT) algorithm to map $\mathcal{A}^{'}(F_{spa})$ and $\mathcal{P}(F_{spa})$ back to their image space and obtain the frequency features $F_{fre}$. Finally, we adopt channel-wise concatenation followed with one $1\times1$ convolution to fully integrate the cross-domain features $F_{spa}$ and $F_{fre}$. The RPB is similar to RAB, just swapping the operations between $\mathcal{A}$ and $\mathcal{P}$.
	
	\subsection{Adaptive Triple Interaction Module}
	The information exchange is an important ingredient in the two-stage design because simply passing the previous stage's output to the next stage neglects the potential intermediate features, but the original, useful shallow features will be gradually weakened with the increasing network depths. In our framework, we construct the ATIM to associate two stages by adequately incorporating cross-domain, cross-scale, and cross-stage features.  
	
	\textbf{\textit{As for the cross-domain interaction}}, the proposed RAB and RPB employed in two stages fully utilize the spatial and frequency domain information to reconstruct the representation of low-frequency (contrast, illumination, and color) and structural components. 
	
	\textbf{\textit{As for the cross-scale interaction}}, in the first stage's decoder and second stage's encoder, the ATIM integrates information from all scales, allowing for both top-down and bottom-up information flow. Further, this procedure combines features with different receptive fields, which enriches contextual representations. As shown in Figure~\ref{fig:block} (b), the features $\big\{D_{1}^{si}\big\}_{i=1}^3$ in the decoder of the first stage firstly will be extracted by three independent $3\times3$ convolutions. Upsampling and downsampling operations will further resize the obtained resolution features. To achieve the combination of precisely spatial high-resolution features with low-resolution features containing rich contextual information, a simple channel concatenation operation followed by $1\times1$ convolution is used to fuse them. The resultant multi-scale features can be denoted as $\big\{\widehat{D}_{1}^{si}\big\}_{i=1}^3$. Similarly, one can obtain the fused features $\big\{\widehat{E}_{2}^{si}\big\}_{i=1}^3$ in the encoder of the second stage. 
	
	\textbf{\textit{As for the cross-stage interaction}}, the ATIM is equipped with an adaptive dynamic filter block (ADFB), which generates weight filters with flexibility based on the fused cross-domain and cross-scale features and applies them to the decoder's features of the second stage for representation capability enhancement. Taking the specific-scale feature set $\big\{\widehat{D}_{1}^{s1},\widehat{E}_{2}^{s1}, D_{2}^{s1} \big\} \in R^{c\times h \times w}$ as an example. The ADFB first uses one $1\times1$ convolution to fuse three inputs and then embeds resulting features into lightweight spatial and channel context branches to enhance filter representations. The spatial context branch includes a $3\times3$ depthwise convolution~\cite{mobilenets}, whereas the channel context branch consists of a pooling layer and a convolutional layer. To take full advantage of them, the element-wise addition operation is used to integrate them. In order to formulate spatially-varying filters, a $1\times1$ convolutional layer is utilized to produce feature maps whose dimension is $(k\times k \times c)\times h \times w$; we then reshape them into a series of per-pixel kernels $W_{i,j} \in \mathbb{R}^{k^2 \times c}$, where $i \in \{1,2,\cdots,h\}$, $j \in \{1, 2, \cdots, w\}$. In this way, a series of content-adaptive filters for each location is learned. Finally, given the scale-specific features $D_{2}^{s1}$, the dynamic filter results can be written as: 
	\begin{equation}
		\widehat{D}_{2}^{s1} = D_{2}^{s1} \ast W + D_{2}^{s1}, 
	\end{equation}
	where ``$\ast$'' indicates the convolution operation, $\widehat{D}_{2}^{s1}$ is the enhanced features in the decoder of the second stage. Here,  we adopt residual learning to preserve more low-level features.
	
	Here, we mainly emphasize two points. (1) The generated filters of ADFB are conditioned on the triple interaction features, thus being guided by the enriched spatial-frequency features (cross-domain), multi-scale contextual features (cross-scale), and deep-shallow features (cross-stage). Such a mechanism allows the ADFB to adapt flexibly to image contents. (2) The two stages explore different frequency domain characteristics and extract rich spatial contextual features, so redundant features still exist. 
	
	\subsection{Mutual Information Constraint}
	Given the multi-scale features $\big\{D_{1}^{si}\big\}_{i=1}^3$ from the first stage's decoder and $\big\{E_{2}^{si}\big\}_{i=1}^3$ from the second stage's encoder, the first step is to transform them into low-dimensional feature vectors for feature embedding. In detail, the paired features of the same scale are inputted into two $3\times3$ convolutional layers. Then the obtained features are mapped using two fully connected layers (``FC'' in Figure~\ref{fig:net}) to produce a low-dimensional feature vector set $\big\{v_d^{si}\big\}_{i=1}^3$ and $\big\{v_e^{si}\big\}_{i=1}^3$. After obtaining the embedded vectors $v_{d}$ and $v_{e}$, the next step is to introduce the mutual information minimization constraint to mitigate feature redundancy and achieve the complementary learning of diverse features.
	
    According to the information theory~\cite{kraskov2004estimating}, the mutual information between $v_{d}$ and $v_{e}$, with the joint entropy $G\left(v_d | v_e\right)$, and marginal entropies $G\left(v_d\right)$ and $G\left(v_e\right)$, can be expressed as:
	\begin{equation}
		I\left(v_d, v_e\right)=G\left(v_d\right)+G\left(v_e\right)-G\left(v_d | v_e\right).
		\label{mi_eq_1}
	\end{equation}
    % where $G(\cdot)$ represents the entropy, $G\left(v_d\right)$ and $G\left(v_e\right)$ are the marginal entropies, and the $G\left(v_d | v_e\right)$ is the joint entropy of $v_{d}$ and $v_{e}$. 
    Formally, following~\cite{info3d}, the representation of KL divergence of these two variables is also given as:
	\begin{equation}
		\begin{aligned}
			& K\left(v_d \| v_e\right)=G_{v_e}\left(v_d \right)-G\left(v_d\right), \\
			& K\left(v_e \|z_d\right)=G_{v_d}\left(v_e\right)-G\left(v_e\right),
		\end{aligned}
		\label{mi_eq_2}
	\end{equation}
	where $G_{v_e}\left(v_d \right)$ and $G_{v_d}\left(v_e \right)$ represent the cross-entropy. Thereby, combining Eq.~\ref{mi_eq_1} and Eq.~\ref{mi_eq_2}, we can obtain:
	\begin{equation}
		\begin{aligned}
			I\left(v_d, v_e\right) = & G_{v_d}\left(v_e\right)+G_{v_e}\left(v_d\right)- G\left(v_d|v_e\right) \\ & -K\left(v_d \| v_e\right)-K\left(v_e \| v_d\right).
		\end{aligned}
	\end{equation}
	Due to the non-negative property of $G\left(v_d | v_e\right)$, minimizing the mutual information is equivalent to optimizing the following formula:
	\begin{equation}
		\begin{aligned}
			L_{mi} = G_{v_d}\left(v_e\right)+G_{v_e}\left(v_d\right) -K\left(v_d \| v_e\right) - K\left(v_e \| v_d\right).
		\end{aligned}
	\end{equation}
	To enforce the mutual information constraint on paired scale-specific features, we use a multi-scale joint learning manner. By minimizing $I(v_e, v_d)$, one can explore the complementary characteristics of two stages effectively. Additionally, the MIC is only used in the training phase and thus does not affect the model's efficiency.
	\subsection{Loss Function}
	Since there are two outputs, $Y_1$ and $Y_2$, from our two-stage network, then the loss function $L_1$ and $L_2$ for both stages is:
	%\begin{small}
	\begin{equation}
		L_{1}=\underbrace{\left\|Y_{1}-\mathcal{F}^{-1}(\mathcal{A}(X_{gt}), \mathcal{P}(X_{hazy}))\right\|_1}_{spatial} + \alpha \underbrace{\left\|\mathcal{A}\left(Y_{1}\right)-\mathcal{A}\left(X_{gt }\right)\right\|_1}_{frequency}, 
	\end{equation}
	%\end{small}
	%\begin{small}
	\begin{equation}
		L_{2}=\underbrace{\left\|Y_2-X_{gt}\right\|_1}_{spatial} + \beta \underbrace{\left\|\mathcal{F}\left(Y_2\right)-\mathcal{F}\left(X_{gt}\right)\right\|_1}_{frequency}.
	\end{equation}
	%\end{small}
	Both terms in each equation above are performed on spatial and frequency domains, respectively. $\left\| \cdot \right\|$ denotes the Mean Absolute Error (MAE), $\alpha$ and $\beta$ are the trade-off factor, and we set them as 0.05. As a consequence, the overall loss after combining with mutual information minimization loss is:
	\begin{equation}
		L = L_{1} + L_{2} + \gamma \sum_{i=1}^3 L_{mi}\left(v_d^{si}, v_e^{si}\right).
	\end{equation}
	As the $L_{mi}$ is much larger than other terms, we empirically set the weight $\gamma=0.001$ for balanced learning. 
	
	\section{Experiments}
	We will first describe the experimental setup in this section. Then, experimental results and a detailed ablation analysis of our proposed methods will be illustrated. 
	
	\begin{figure*}[htbp]
		\centering
		\begin{minipage}{\textwidth}		   					
			\centering
			\subfigure[]{\includegraphics[width=.135\linewidth]{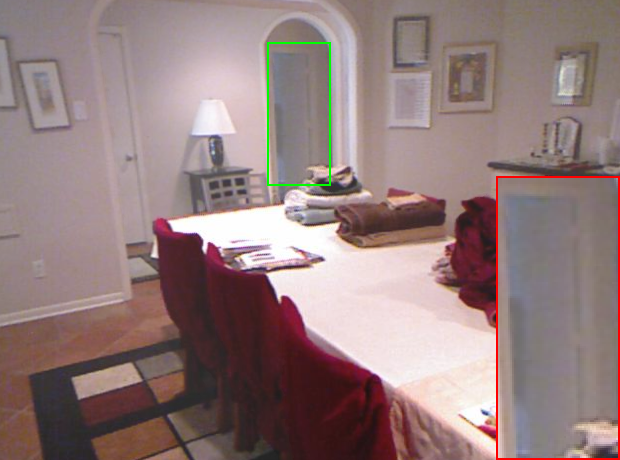}}
			\subfigure[]{\includegraphics[width=.135\linewidth]{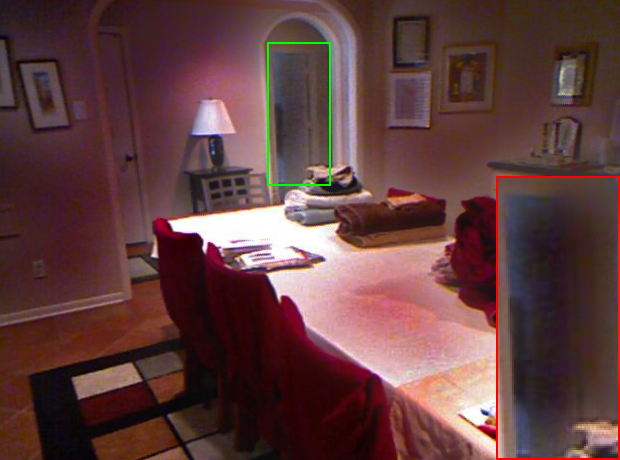}}
			\subfigure[]{\includegraphics[width=.135\linewidth]{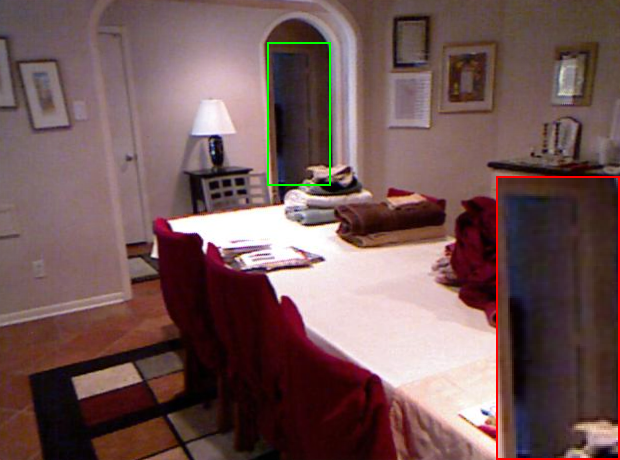}}
			\subfigure[]{\includegraphics[width=.135\linewidth]{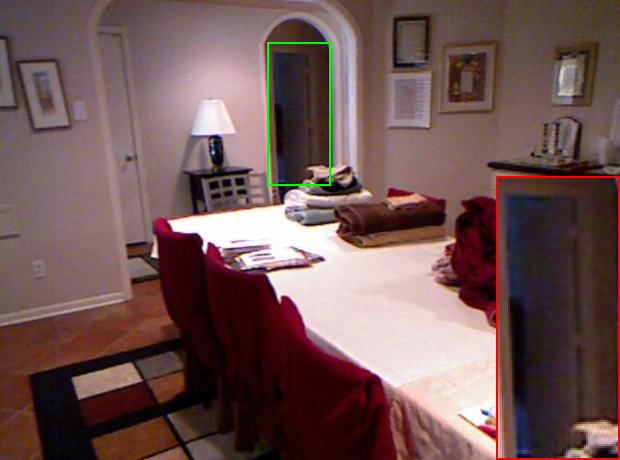}} 
			\subfigure[]{\includegraphics[width=.135\linewidth]{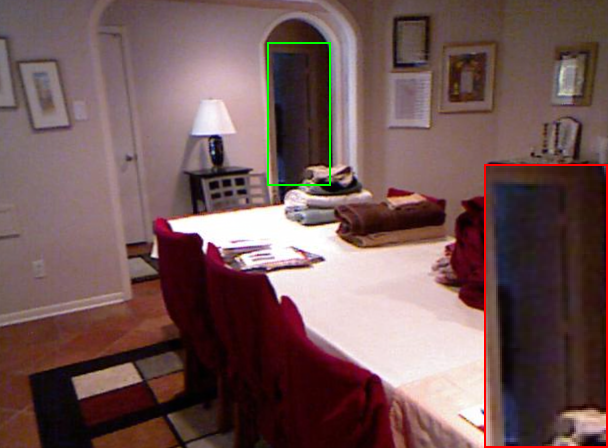}} 
			\subfigure[]{\includegraphics[width=.135\linewidth]{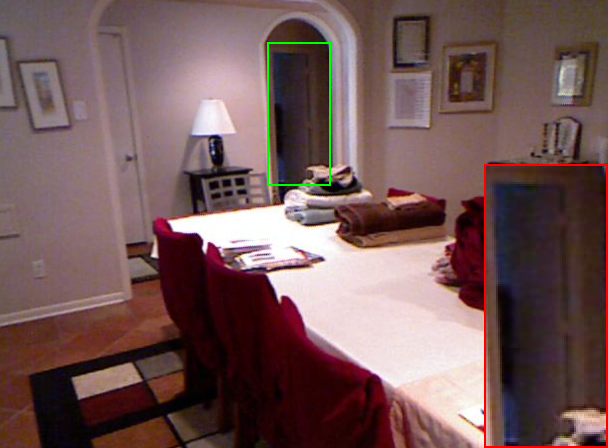}}
			\subfigure[]{\includegraphics[width=.135\linewidth]{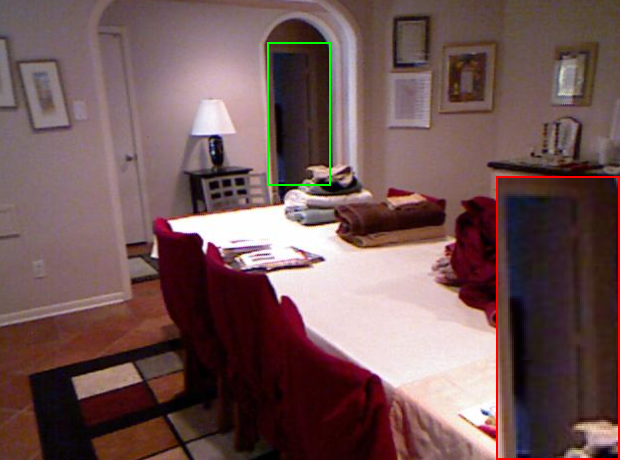}}
		\end{minipage}
		\begin{minipage}{\textwidth}		   					
			\centering
			\vspace{-0.6cm}
			\subfigure[(a) Hazy]{\includegraphics[width=.135\linewidth]{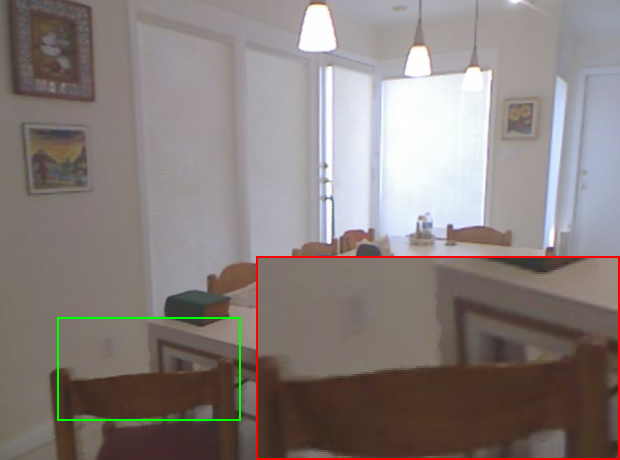}}
			\subfigure[(b) DCP~\cite{dcp}]{\includegraphics[width=.135\linewidth]{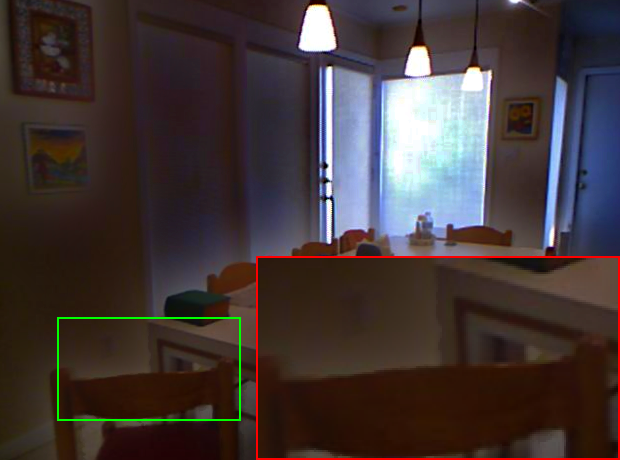}}
			\subfigure[(c) FFA-Net~\cite{ffa_net}]{\includegraphics[width=.135\linewidth]{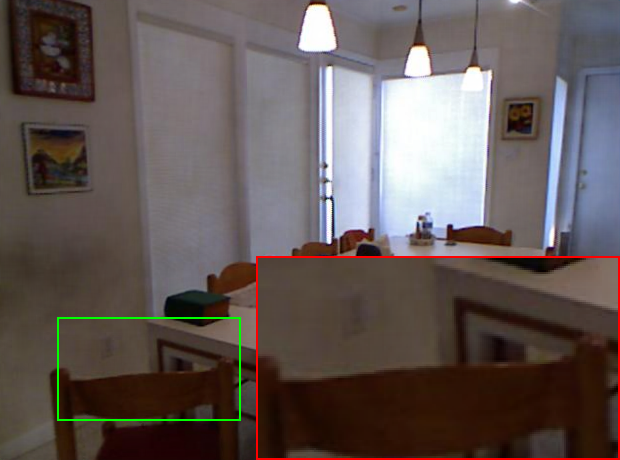}}
			\subfigure[(d) AECR-Net~\cite{aecr_net}]{\includegraphics[width=.135\linewidth]{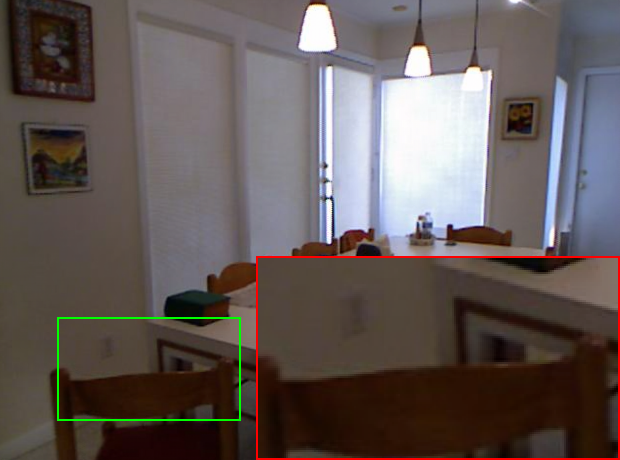}} 
			\subfigure[(e) Dehamer~\cite{dehamer}]{\includegraphics[width=.135\linewidth]{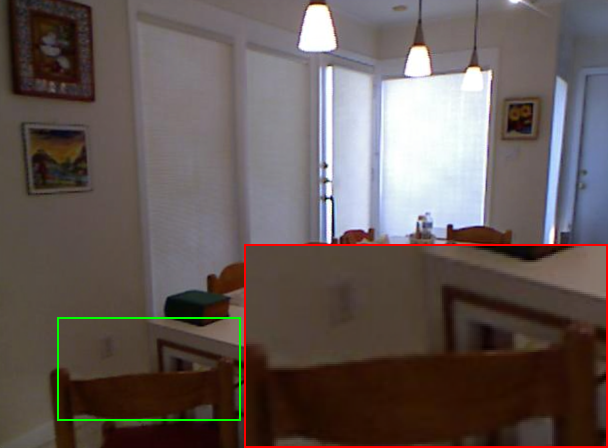}} 
			\subfigure[(f) \textbf{MITNet (Ours)}]{\includegraphics[width=.135\linewidth]{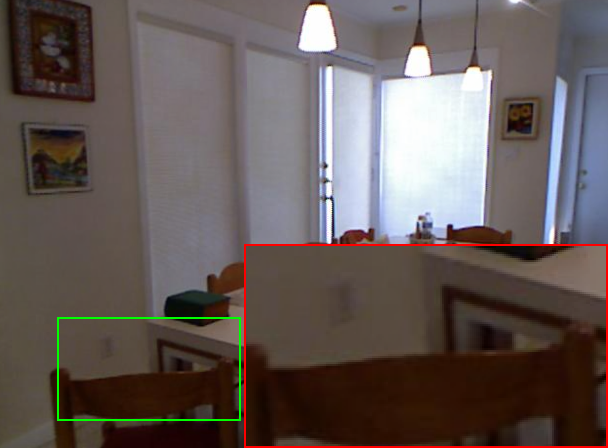}}
			\subfigure[(g) GT]{\includegraphics[width=.135\linewidth]{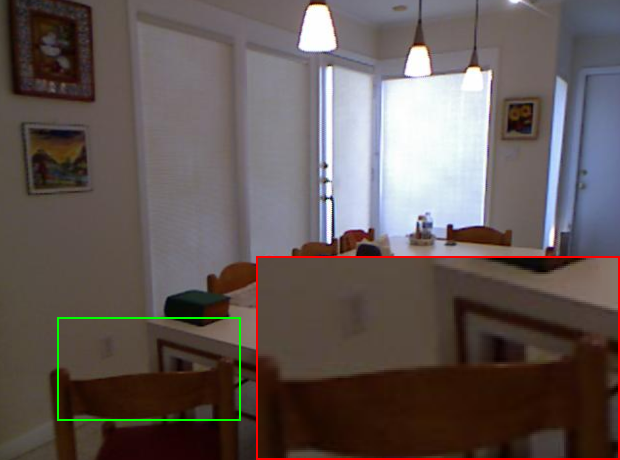}}
		\end{minipage}
		\vspace{-2mm}
		\caption{Visual comparisons of SOTS~\cite{reside} dataset by different methods.}
		\vspace{-2mm}
		\label{fig:sots}
	\end{figure*}

	\subsection{Experiment Setting}
	\textbf{\textit{Datasets.}} For fair comparisons with existing dehazing methods, we utilize \textbf{ITS} and \textbf{OTS} subsets of RESIDE~\cite{reside} dataset as the training data and evaluate performance on \textbf{SOTS} subset for synthetic image dehazing. For real image dehazing, two common real-world datasets, \textbf{Dense-Haze}~\cite{dense_haze} and \textbf{NH-Haze}~\cite{nh_haze}, are adopted to test the robustness of our MITNet. It should be emphasized that we execute all evaluations on these datasets separately and do not use extra data to boost results.
	
	\noindent 
	\textbf{\textit{Implementation Details.}}  From the first to the fourth scale, the encoder and decoder have 20, 40, 80, and 160 channels, respectively. The upsampling and downsampling layers are implemented by transposed and strided convolution, respectively. Our model is implemented using the \textit{PyTorch} library on two NVIDIA GeForce GTX 2080Ti GPUs. ADAM optimizer with $\beta_{1}=0.9$, $\beta_{2} = 0.999$, $\epsilon=10^{-8}$ is used in the whole training phase. We set the initial learning rate as 0.0002 and then linearly decrease it to half every $200$ epoch. In addition, 16 patches of size 256$\times$256 are cropped randomly as input images for each training. And in each mini-batch, we enrich and augment these patches to expand training samples by flipping horizontally or vertically and rotating \ang{90}.  
	
	\noindent
	\textbf{\textit{Comparison Methods and Evaluation Metrics.}} We compare our MITNet with four earlier method including DCP~\cite{dcp}, DehazeNet~\cite{dehazenet}, AOD-Net~\cite{aod_net}, GFN~\cite{gfn} and eight recent competing methods including FFA-Net~\cite{ffa_net}, MSBDN~\cite{msbdn}, AECR-Net~\cite{aecr_net}, UDN~\cite{udn}, PMDNet~\cite{pmdnet}, Dehamer~\cite{dehamer}, MAXIM~\cite{maxim}, and FSDGN~\cite{fsdgn}. Additionally, all of the compared results are provided by the authors or retrained by their  available codes(\eg~the real-world dehazing results on FSDGN~\cite{fsdgn}). Note that \textbf{(1)} we do not compare recent methods (\eg~ DehazeFormer~\cite{DehazeFormer} and SFNet\cite{SFNet}) with very high computational cost and network parameters; \textbf{(2)} AECR-Net~\cite{aecr_net} uses the last five images in the real-world training set for evaluation, leading to the reported results being inconsistent with the testing set; we utilize their provided pre-trained models to obtain the results on two testing sets. In addition, Peak Signal-to-Noise Ratio (PSNR) and Structural Similarity Index (SSIM)~\cite{ssim} are utilized for dehazing performance evaluation.
	
	\subsection{Comparison with State-of-the-art Methods}
	In this subsection, the results of several compared methods with ours on synthetic and real-world dehazing datasets are provided.
	
	\textbf{\textit{Evaluation on Synthetic Dataset.}} The quantitative evaluation of our approach and other methods on the SOTS dataset are listed in Table~\ref{table:sys_data}. As observed, our MITNet acquires the highest PSNR on the SOTS-indoor dataset, the second highest PSNR on the SOTS-outdoor dataset, and outperforms all other methods on the SSIM index. Specifically, compared to the Transformer-based method Dehamer and spatial-frequency guided method FSDGN, we achieve performance gains by 3.6 dB and 1.6 dB on the SOTS-indoor dataset. Although there is no performance improvement on the SOTS-outdoor compared to Dehamer~\cite{dehamer}, our method is more lightweight and efficient. We also show the visualized comparisons with representative methods on the SOTS dataset in Figure~\ref{fig:sots}, from which we can see that the compared methods retain haze or produce color deviations. In contrast, our method preserves more details and involves fewer color distortions, resulting in the closest match to the original clean images and accurate haze removal.

	\begin{table}[t]
		\caption{Quantitative comparisons of various dehazing methods on SOTS (indoor and outdoor)~\cite{reside}.}
		% \vspace{-2mm}
		\label{table:sys_data}
		\centering
		\scalebox{.9}{
			\begin{tabular}{|l |c ||c c |c c |}
				\hline 
				\rowcolor{mygray}
				&
				&\multicolumn{2}{c|}{SOTS-indoor}
				&\multicolumn{2}{c|}{SOTS-outdoor}\\
				\cline{3-6}
				\rowcolor{mygray}
				\multirow{-2}*{Method}
				&\multirow{-2}*{Venue}
				& PSNR~$\textcolor{black}{\uparrow}$ & SSIM~$\textcolor{black}{\uparrow}$ 
				& PSNR~$\textcolor{black}{\uparrow}$ & SSIM~$\textcolor{black}{\uparrow}$ \\
				\hline \hline
				DCP~\cite{dcp}                      & TPAMI'10   & 16.61	 & 0.8546	 & 19.14	& 0.8605       \\ 
				DehazeNet~\cite{dehazenet}          & TIP'16     & 19.82    & 0.8209    & 27.75    & 0.9269       \\ 
				AOD-Net~\cite{aod_net}              & ICCV'17    & 20.51    & 0.8162    & 24.14    & 0.9198                 \\ 
				GFN~\cite{gfn}                      & CVPR'18    & 22.30    & 0.8800    & 21.55    & 0.8444                 \\ \hline
				FFA-Net~\cite{ffa_net}              & AAAI'20    & 36.39    & 0.9886    & 33.57    & 0.9840                 \\ 
				MSBDN~\cite{msbdn}                  & CVPR'20    & 32.77    & 0.9812    & 34.81    & 0.9857                 \\ 
				AECR-Net~\cite{aecr_net}            & CVPR'21    & 37.17    & 0.9901    & -        & -                      \\ 
				UDN~\cite{udn}                      & AAAI'22    & 38.62    & 0.9909    & 34.92    & 0.9871                    \\
				PMDNet~\cite{pmdnet}                & ECCV'22    & 38.41    & 0.9900    & 34.74    & 0.9850              \\ 
				Dehamer~\cite{dehamer}              & CVPR'22    & 36.63    & 0.9881    & \textbf{35.18}    & 0.9860            \\ 
				MAXIM-2S~\cite{maxim}               & CVPR'22    & 38.11    & 0.9908    & 34.19    & 0.9846                \\
				FSDGN~\cite{fsdgn}                  & ECCV'22    & 38.63    & 0.9903    & -        & -                \\ \hline
				%				DehazeFormer             & TIP'23     & 
				\textbf{MITNet (Ours)}            & -          & \textbf{40.23}    & \textbf{0.9920}    & \textbf{35.18}    & \textbf{0.9881}       \\ \hline
		\end{tabular}}		
		\vspace{-2mm}
	\end{table}
	
	\begin{table}[t]
		\begin{center}
			\caption{Quantitative comparisons of various dehazing methods on Dense-Haze~\cite{dense_haze} and NH-HAZE~\cite{nh_haze}.}
			% \vspace{-2mm}
			\label{table:real_data}
			\scalebox{.9}{
				\begin{tabular}{|l |c ||c c| c c|}
					\hline
					\rowcolor{mygray}
					&
					&\multicolumn{2}{c}{Dense-Haze}
					&\multicolumn{2}{c|}{NH-HAZE}\\ 
					\cline{3-6}
					\rowcolor{mygray}
					\multirow{-2}*{Method}
					& \multirow{-2}*{Venue}
					& PSNR~$\textcolor{black}{\uparrow}$ & SSIM~$\textcolor{black}{\uparrow}$ 
					& PSNR~$\textcolor{black}{\uparrow}$ & SSIM~$\textcolor{black}{\uparrow}$ \\
					\hline \hline
					DCP~\cite{dcp}                      & TPAMI'10   & 10.06	 & 0.3854	 & 10.57	& 0.5196     \\ 
					DehazeNet~\cite{dehazenet}          & TIP'16     & 13.84    & 0.4252    & 16.62    & 0.5238     \\ 
					AOD-Net~\cite{aod_net}              & ICCV'17    & 13.14    & 0.4144    & 15.40    & 0.5693     \\ 
					GDN~\cite{gdn}                      & ICCV'19    & 13.31    & 0.3681    & 13.80    & 0.5370     \\ \hline
					FFA-Net~\cite{ffa_net}              & AAAI'20    & 14.39    & 0.4524    & 19.87    & 0.6915     \\ 
					MSBDN~\cite{ffa_net}                & CVPR'20    & 15.37    & 0.4858    & 19.23    & 0.7056     \\ 
					AECR-Net~\cite{aecr_net}            & CVPR'21    & 14.88    & 0.5049    & 19.92    & 0.6717     \\ 
					%				PMDNet                   & ECCV'22    & 16.79    & 0.5100    & 20.42    & 0.7300     \\ 
					Dehamer~\cite{dehamer}              & CVPR'22    & 16.62    & 0.5602    & 20.66    & 0.6844      \\ 
					FSDGN~\cite{fsdgn}                  & ECCV'22    & 16.91    & 0.5806    & 19.99    & 0.7086     \\ \hline
					%				DehazeFormer             & TIP'23     & 
					\textbf{MITNet (Ours)}            & -          & \textbf{16.97}    & \textbf{0.6056}    & \textbf{21.26}         & \textbf{0.7122}          \\ \hline
			\end{tabular}}
		\end{center}
	\end{table}
	
	\begin{figure*}[t]
		\begin{center}
			\begin{minipage}{\textwidth}
				\centering
				\vspace{0.05cm}
				\includegraphics[width=0.15\linewidth]{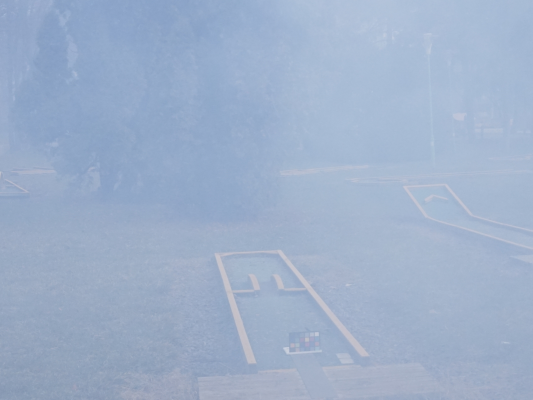} 
				\includegraphics[width=0.15\linewidth]{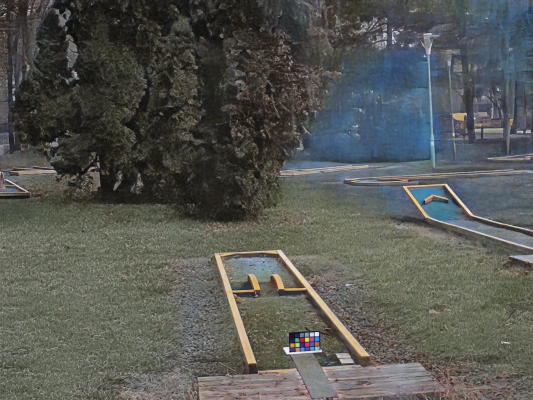}
				\includegraphics[width=0.15\linewidth]{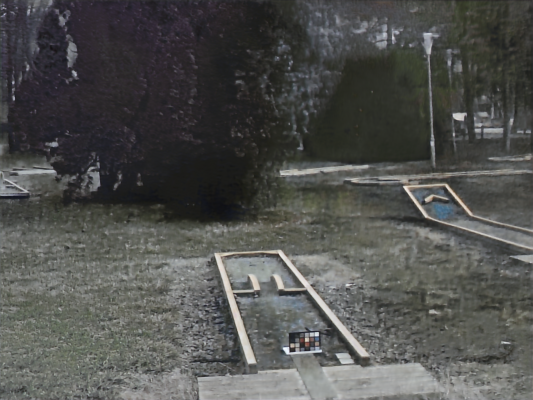}
				\includegraphics[width=0.15\linewidth]{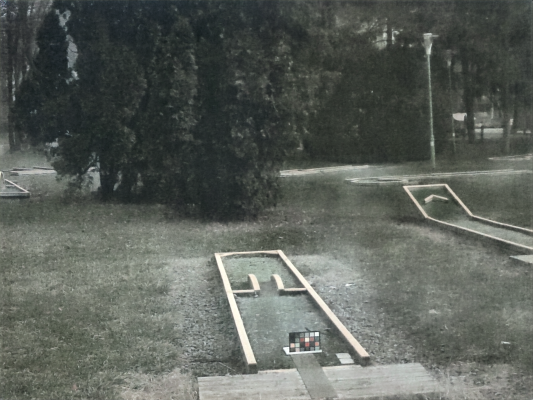}
				\includegraphics[width=0.15\linewidth]{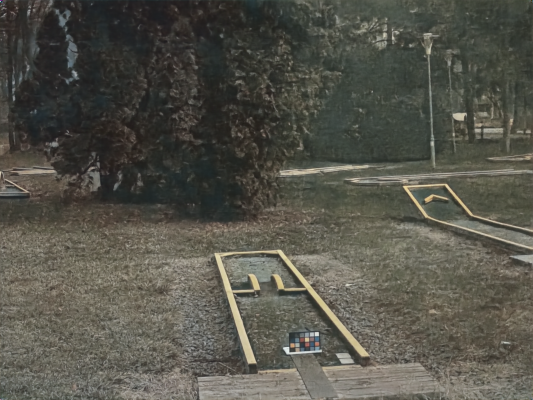}
				\includegraphics[width=0.15\linewidth]{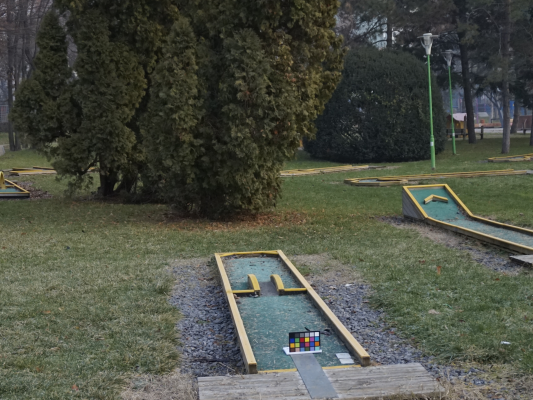}
			\end{minipage}
			\begin{minipage}{\textwidth}
				\centering
				\vspace{-0.1cm}
				\subfigure[(a) Hazy]{\includegraphics[width=0.15\linewidth]{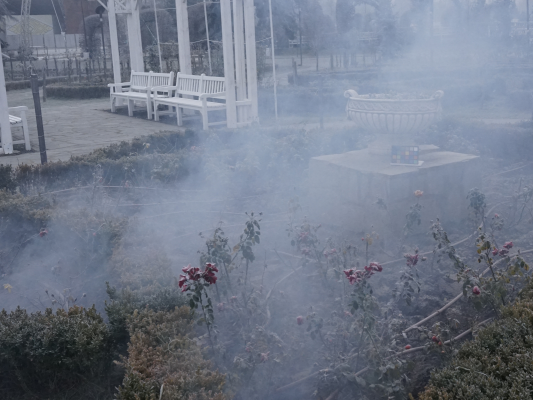}}
				\subfigure[(b) AECR-Net~\cite{aecr_net}]{\includegraphics[width=0.15\linewidth]{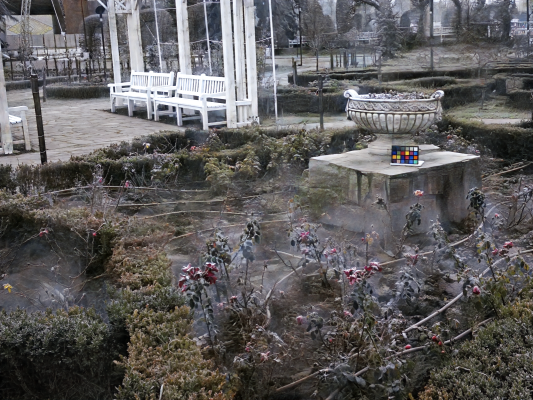}}
				\subfigure[(c) Dehamer~\cite{dehamer}]{\includegraphics[width=0.15\linewidth]{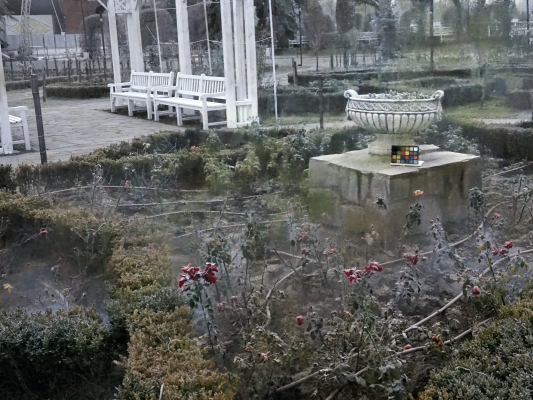}}
				\subfigure[(d) FSDGN~\cite{fsdgn}]{\includegraphics[width=0.15\linewidth]{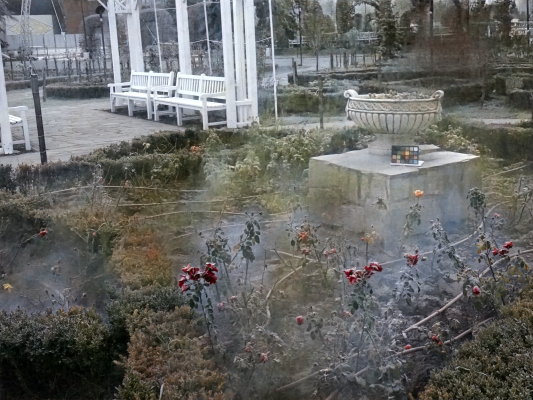}}
				\subfigure[(e) \textbf{MITNet (Ours)}]{\includegraphics[width=0.15\linewidth]{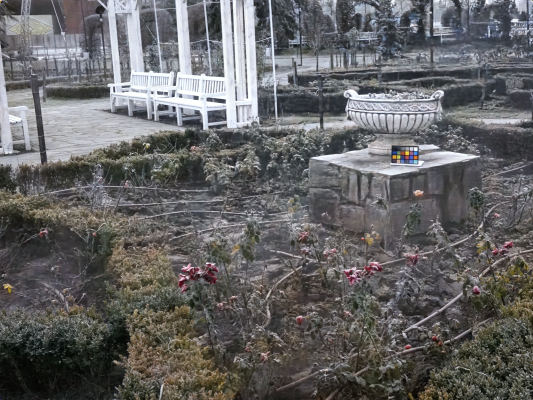}}
				\subfigure[(f) GT]{\includegraphics[width=0.15\linewidth]{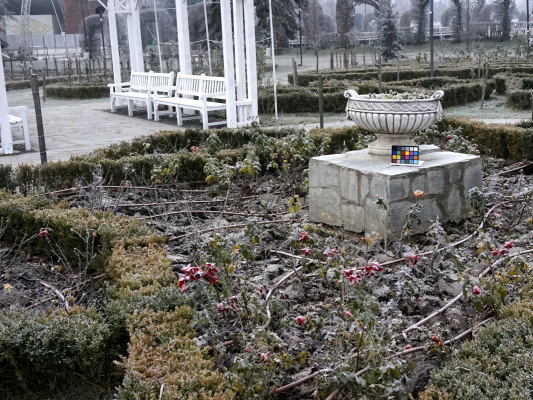}}
			\end{minipage}
		\end{center}
		\vspace{-2mm}
		\caption{Visual comparisons of real-world image dehazing methods on Dense-Haze~\cite{dense_haze} and NH-Haze~\cite{nh_haze} datasets.}
		\vspace{-2mm}
		\label{fig:real}
	\end{figure*}
	
	\begin{figure*}[htbp]
		\centering
		\includegraphics[width=0.3\linewidth]{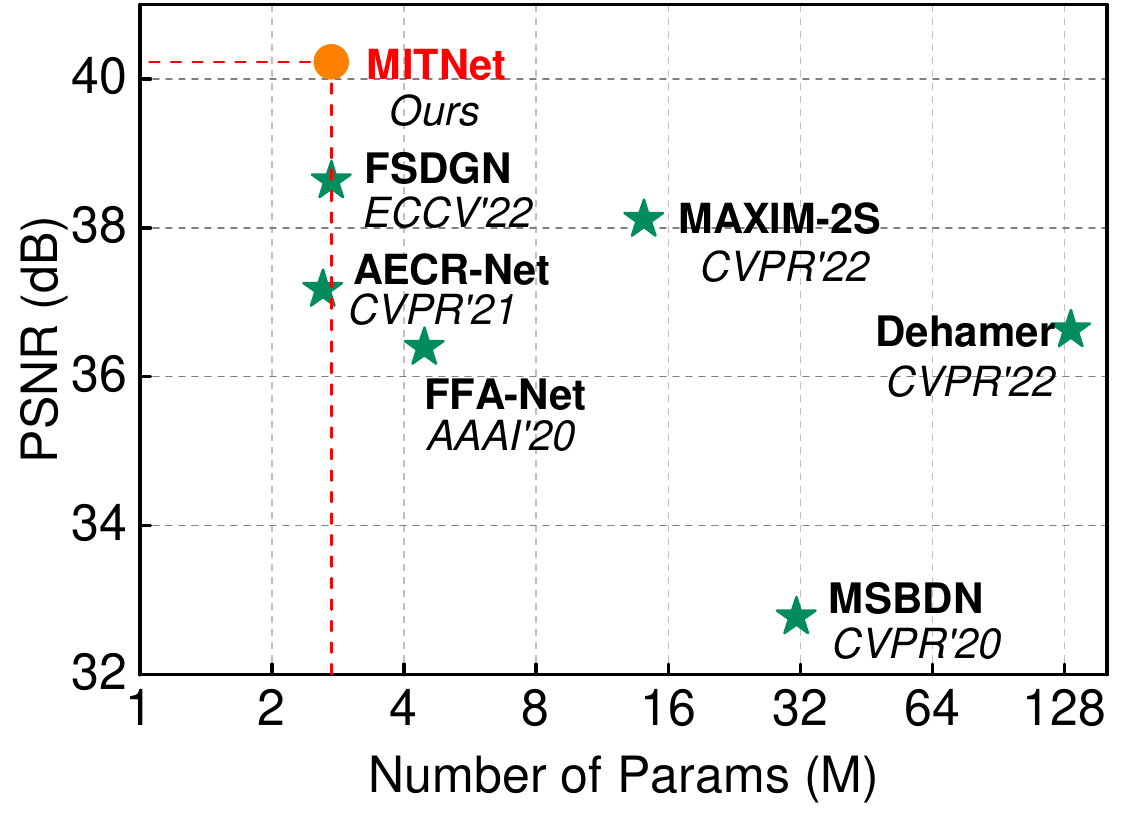}
		\includegraphics[width=0.3\linewidth]{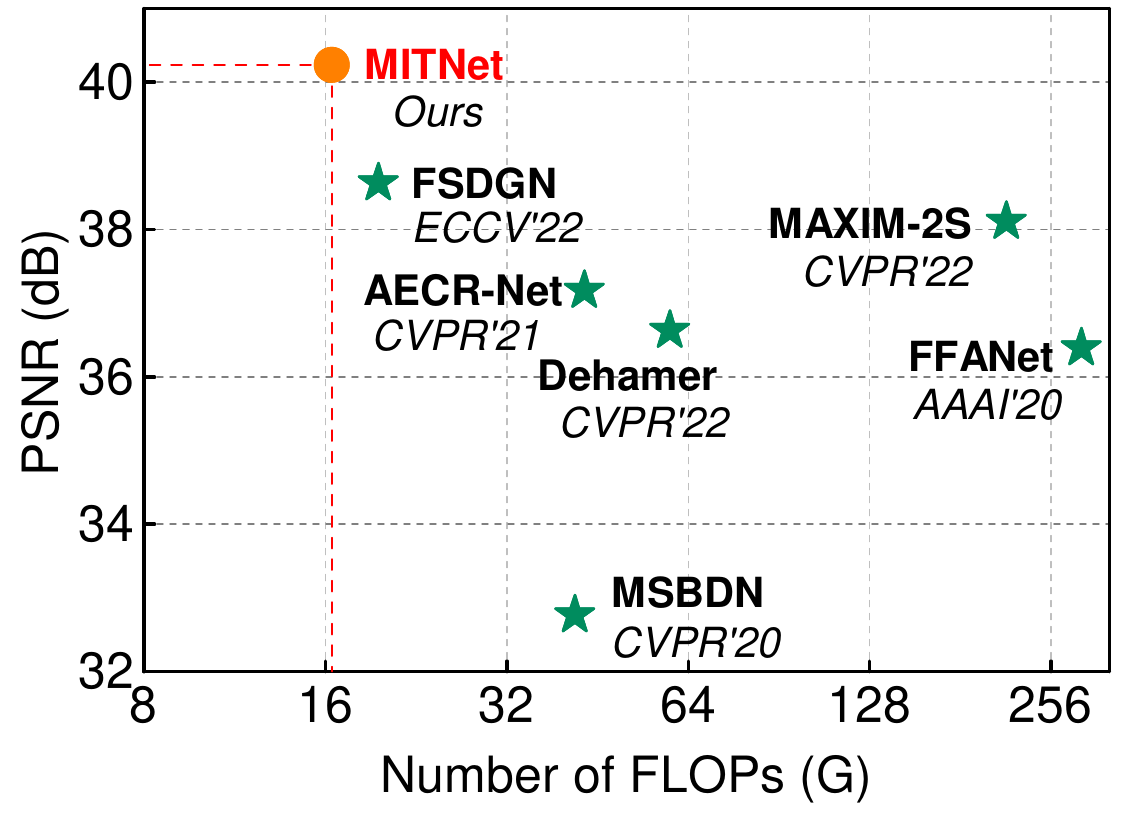}
		\includegraphics[width=0.3\linewidth]{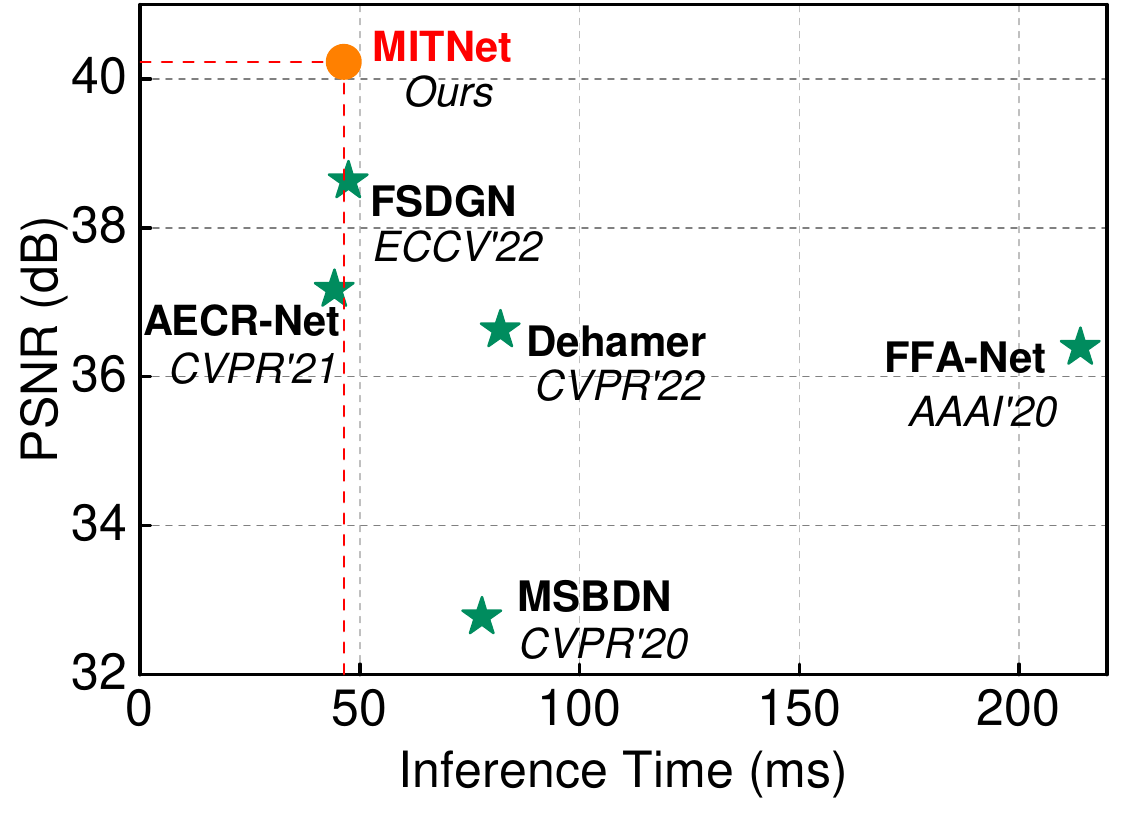}		
		\vspace{-3mm}
		\caption{Comparisons of model complexity of different methods. FLOPs are calculated with an input size of 256$\times$256, and the average inference time is evaluated on SOTS-indoor~\cite{reside}.}
		\vspace{-3mm}
		\label{fig:complexity}
	\end{figure*}
	
	\textbf{\textit{Evaluation on Real-world Datasets.}} We also perform the evaluation on real-world datasets (\eg~ the Dense-Haze and NH-Haze) and present the quantitative results in Table~\ref{table:real_data}. We all know that removing haze from real-world photographs is significantly more challenging than removing haze from synthetic photographs due to the dense and nonhomogeneous haze distribution. However, the proposed MITNet surpasses all compared methods on both datasets. The visually compared images are also presented in Figure~\ref{fig:real}. We can see although other competitors successfully remove most of the haze, their results lose the colorfulness of images. Our method effectively removes the homogeneous or nonhomogeneous haze and reconstructs vivid colors. 

	\subsection{Model Complexity Analyses}
	We further evaluate the model complexity (\eg~network parameters, FLOPs, and inference time) of deep learning-based dehazing approaches over the last three years. The average running time is evaluated on the SOTS-indoor dataset, FLOPs are obtained based on 256$\times$256 resolution image patch, and each result is acquired by repeating the experiment five times to guarantee a fair comparison. From Figure~\ref{fig:complexity}, we can easily conclude that our MITNet obtains the best performance with the fewest number of parameters, the fewest FLOPs, and the second-fast inference time. Thus, our MITNet achieves an excellent model complexity versus performance trade-off. FSDGN~\cite{fsdgn} utilizes dual-domain information while having a noticeable performance gap with ours, which shows it is necessary for adequate feature interaction and redundant feature removal.
	
	\begin{figure*}[t]
		\begin{center}
			\includegraphics[width=0.94\linewidth]{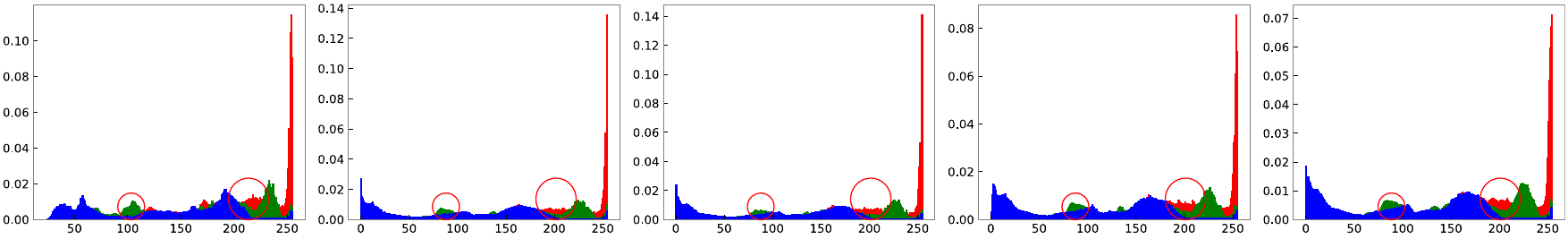}
			\includegraphics[width=0.94\linewidth]{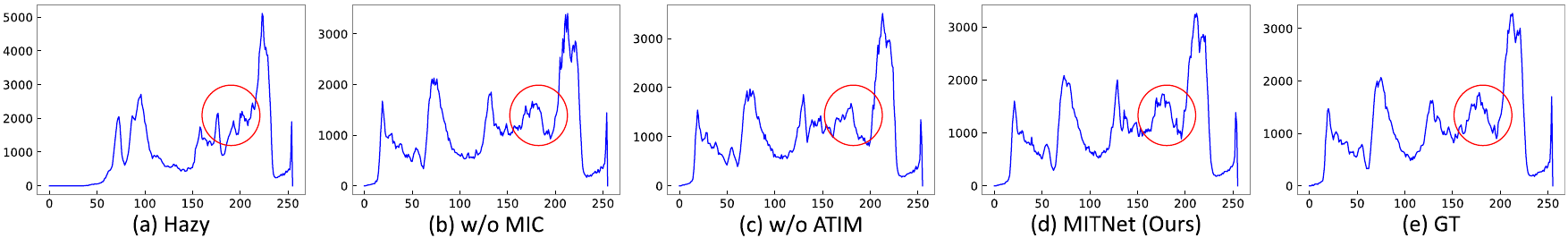}
		\end{center}
		\vspace{-3mm}
		\caption{Visualization of the histograms of various images. (a) Input hazy images, (b) output of MITNet without mutual information constraint (w/o MIC), (c) output of MITNet without adaptive triple interaction module (w/o ATIM), (d) output of our MITNet, and (e) ground truth images.}
		\vspace{-3mm}
		\label{fig:hist}
	\end{figure*}

	\subsection{Ablation Studies}
	% We conduct ablation studies to verify further our proposed two-stage architecture design, adaptive triple interaction module, and mutual information constraint.

        \begin{table}[t]
	\caption{Ablation study about the two-stage design and the combination of spatial-frequency domain features.}	
        \vspace{-2mm}
        \label{table:two_stage}
		\scalebox{0.83}{
			\begin{tabular}{|l ||c | c| c| c| c| c|}
				\hline
				\rowcolor{mygray}
				Model 
				& $M_{1}$
				& $M_{2}$
				& $M_{3}$
				& $M_{4}$
				& $M_{5}$
				& $M_{6}$ \\ \hline \hline
				\#Parmas(M)  & 2.08    & 2.22   & 2.22	   & 2.39      & 2.39   & 2.39   \\
				PSNR(dB)     & 35.43   & 35.98  & 36.11   & 34.17     & 37.43  & \textbf{37.64}\\
				SSIM         & 0.9870  & 0.9890 & 0.9899  & 0.9863    & 0.9905 & \textbf{0.9908}\\ \hline 
		\end{tabular}}
	\end{table}
	\begin{table}[t]
		\caption{Ablation investigation on the effectiveness of the proposed ATIM and MIC.}
            \vspace{-2mm}
		\label{table:component}		
		\scalebox{0.83}{
			\begin{tabular}{|c||c|c|c|c|c|c|} \hline 
				\rowcolor{mygray}
				Model
				& Triple Interaction 
				& ADFB 
				& MIC
				& \makecell{\#Params\\ (M)}
				& \makecell{PSNR \\ (dB)}
				& SSIM     \\ \hline \hline
				$M_{a}$   & \xmark    & \xmark     & \xmark  & 2.39 & 37.64 & 0.9908  \\  
				$M_{b}$   & \cmark    & \xmark     & \xmark  & 2.54 & 38.64 & 0.9913\\ 
				$M_{c}$   & \xmark    & \cmark     & \xmark  & 2.49 & 38.81 & 0.9910\\ 
				$M_{d}$   & \cmark    & \cmark     & \xmark  & 2.73 & 39.47 & 0.9916\\
				$M_{e}$   & \cmark    & \cmark     & \cmark  & 2.73& \textbf{40.23} & \textbf{0.9920} \\ \hline
		\end{tabular}}
	\end{table}
 
	\textbf{\textit{Two-stage Architecture.}} Here, we design several variants of networks to validate the significance and effectiveness of the two-stage architecture. All ablation studies are conducted without using the proposed ATIM and MIC. $M_{1}$ denotes that the Fourier prior is not used in both stages, that is, the overall process is performed on the spatial domain. $M_{2}$ denotes that the RABs used in the first stage are only performed in the spatial domain. $M_{3}$ denotes that RPBs used in the second stage are only performed in the spatial domain. $M_{4}$ denotes exchanging the order of the first stage and the second stage. $M_{5}$ denotes that the network architecture is consistent with our MITNet, but uses $Y_1$ as the input of the second stage. $M_{6}$ denotes the proposed MITNet without using the ATIM and MIC.
	
	As shown in Table~\ref{table:two_stage}, $M_{4}$ presents the worse performance. In this case, the inputs of these two stages contain amplitude spectrum information, which means the whole process is trained on the degraded hazy images. Both $M_{2}$ and $M_{3}$ exceed the $M_{1}$, showing the effectiveness of spatial-frequency dual domain information integration. $M_{6}$ achieves 0.21 dB PSNR gain compared to $M_6$, demonstrating that keeping the $\mathcal{P}(X_{hazy})$ invariant is vital for the training of the second stage. Simultaneously, this phenomenon reveals that the first stage also utilizes the phase spectrum information to some extent, which provides insight guidance for removing redundant information using mutual information constraint. 
	
	\textbf{\textit{Adaptive Triple Interaction Module.}} To verify the effectiveness of the proposed ATIM, we ablate the use of components ``Triple Interaction'' and ADFB. As shown in Table~\ref{table:component}, one can see from the top three rows that the proposed triple interaction manner and ADFB significantly outperform the baseline with slightly increasing parameters. Further, when these two modules are employed, $M_{d}$ achieves 1.77 dB performance gains in terms of PSNR, which suggests utilizing the interactive cross-domain, cross-scale, and cross-stage features to generate content-adaptive dynamic filters and applying them to subsequent decoder features is beneficial for improving representation capability.
	
	\textbf{\textit{Mutual Information Constraint.}} We also verify the proposed MIC scheme and list the results in the last two rows of Table~\ref{table:component}. Because the MIC is only employed during the training phase, it has no effect on inference speed or model parameters. It is clearly seen that Model (e), the final MITNet, gets the only result over 40dB, meaning that using MIC forces complementary feature learning. 
 
\begin{figure}[t]
		\centering
		\includegraphics[width=0.99\linewidth]{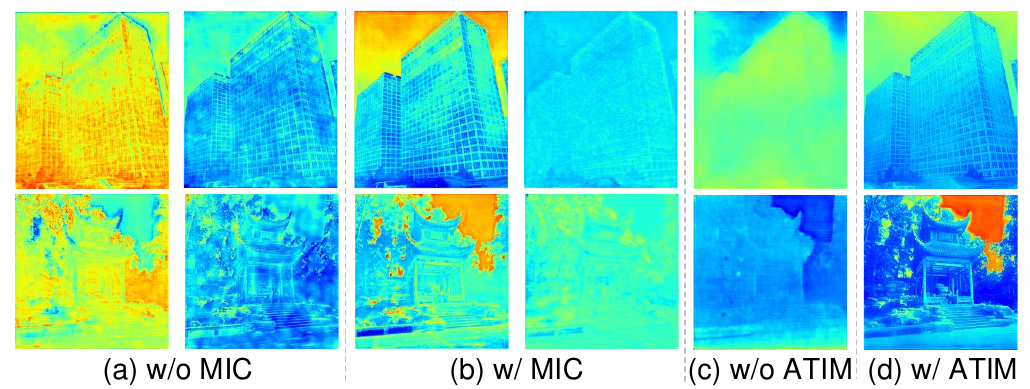}		
		\vspace{-2mm}
		\caption{The features difference between two stages in our model without and with integrating the MIC and ATIM. The first and second columns in (a) and (b) denote features extracted from the first and second stages, respectively.}
		\label{fig:feature}
	\end{figure}

    Figure~\ref{fig:hist} shows the statistical distribution with or without using the proposed modules. The first row is the histogram of the corresponding ablated model's output, representing the illumination and color distribution. The histogram of the corresponding grayscale images is shown in the second row, displaying the variation of texture, light, and contrast. Our MITNet obviously produces a more comparable distribution to the clean photographs. However, the results of ``w/o ATIM'' and ``w/o MIC'' (zoom in on the red circle) have apparent discrepancies. We further show the difference of feature maps in two stages before and after employing MIC and ATIM in Figure~\ref{fig:feature}. We can get some intuitive clues from the visualization. (1) It is apparent that integrating MIC enhances the complementary features learning and reduces the redundant features, thereby producing different textures. (2) The ATIM utilizes the dual domain and spatial-channel contextual information to guide the generation of subsequent features. The resulting feature maps focus on informative structural contents and fine texture details. These experiments indicate that both modules are the key ingredients of our method.
	
	\section{Conclusion}
	In this paper, a two-stage network named MITNet is proposed for image dehazing to address three issues. For the limited investigation of frequency domain information, we propose a two-stage design to construct the network, which progressively restores the haze-free images based on Fourier's amplitude and phase spectrum priors. For the inadequate information interaction, we propose an adaptive triple interaction module to fully integrate cross-domain, cross-scale, and cross-stage features, enhancing model representational capability. For the information redundancy, we utilize the mutual information constraint to learn complementary information and alleviate feature redundancy from two stages. Extensive experiments confirm the validity and generality of our MITNet.
	
	\begin{acks}
		This work was supported in part by the National Natural Science Foundation of China under Grants 61976079 and 62072151; in part by the Key Project of Science and Technology of Guangxi under Grant AB22035022-2021AB20147; and in part by Anhui Provincial Natural Science Fund for the Distinguished Young Scholars under grant 2008085J30.
	\end{acks}
        \clearpage
	\bibliographystyle{ACM-Reference-Format}
	\bibliography{ref}
\end{document}